\documentclass[11pt]{article}

\usepackage[preprint]{acl}
\usepackage{amsmath}
\usepackage{bbold}
\usepackage{times}
\usepackage{amssymb}
\usepackage{latexsym}
\usepackage{booktabs}
\usepackage{graphicx}
\usepackage[table]{xcolor}
\usepackage{xspace}
\usepackage[T1]{fontenc}
\usepackage{xcolor}
\usepackage[most]{tcolorbox}
\newtcblisting{promptbox}[1][]{%
  enhanced,
  breakable,
  colback=gray!10,
  colframe=gray!50,
  arc=1mm,
  boxsep=4pt,
  fontupper=\ttfamily\small,
  title=#1,
  listing only,
  listing engine=listings,
  listing options={
    inputencoding=latin1,    
    breaklines=true,
    breakatwhitespace=false,
    basicstyle=\ttfamily\small,
    columns=fullflexible,
    escapeinside={(*@}{@*)}      
  }%
}

\usepackage[utf8]{inputenc}

\usepackage{microtype}

\usepackage{inconsolata}

\usepackage{graphicx}
\newcommand{\method}{\textit{RelayLLM}\xspace}

\title{\method: Efficient Reasoning via Collaborative Decoding}

\author{Chengsong Huang\textsuperscript{1} \quad Tong Zheng\textsuperscript{2} \quad Langlin Huang\textsuperscript{1} \\
\textbf{Jinyuan Li\textsuperscript{1} \quad Haolin Liu\textsuperscript{3}\quad Jiaxin Huang\textsuperscript{1}}\\[0.5ex] \textsuperscript{1}Washington University in St. Louis  \quad \textsuperscript{2}University of Maryland \quad \textsuperscript{3}University of Virginia \quad \\ {\tt\small \{chengsong, h.langlin, ljinyuan, jiaxinh\}@wustl.edu \quad tzheng24@umd.edu \quad srs8rh@virginia.edu} }

\begin{document}
\maketitle

\begin{abstract} Deploying Large Language Models (LLMs) for complex reasoning is often hindered by high computational costs and latency, while resource-efficient Small Language Models (SLMs) typically lack the necessary reasoning capacity. Existing collaborative approaches, such as cascading or routing, operate at a coarse granularity by offloading entire queries to LLMs, resulting in significant computational waste when the SLM is capable of handling the majority of reasoning steps. To address this, we propose \textbf{\method}, a novel framework for efficient reasoning via token-level collaborative decoding. Unlike routers, \method~empowers the SLM to act as an active controller that dynamically invokes the LLM only for critical tokens via a special command, effectively "relaying" the generation process. We introduce a two-stage training framework, including warm-up and Group Relative Policy Optimization (GRPO) to teach the model to balance independence with strategic help-seeking. Empirical results across six benchmarks demonstrate that \method~improves the average accuracy from 42.5\% to 49.52\%, effectively bridging the performance gap between the two models. Notably, this is achieved by invoking the LLM for only \textbf{1.07\%} of the total generated tokens, offering a 98.2\% cost reduction compared to performance-matched random routers. Our code is available at \url{https://github.com/Chengsong-Huang/RelayLLM}.
\end{abstract}

\section{Introduction}

Large Language Models (LLMs) have demonstrated remarkable capabilities in complex reasoning and problem-solving~\citep{comanici2025gemini,yang2025qwen3, achiam2023gpt}. However, their deployment is often constrained by high computational costs and latency. In contrast, Small Language Models (SLM) are resource-efficient options but typically struggle with hard reasoning tasks due to their limited capacity~\citep{kaplan2020scaling}. This trade-off has motivated the development of collaborative systems that combine the reasoning capabilities of LLMs with the efficiency of smaller models~\citep{hakimov2025price}.

Existing approaches to different-sized model collaboration often rely on ``cascading'' or ``routing'' mechanisms, where a router determines the difficulty of a query and directs it to either a small or large model~\citep{ding2024hybrid,hu2024routerbench,ong2024routellm}. While effective to some extent, these methods typically operate at a coarse granularity by offloading the entire generation task to the large model once a query seems difficult. This ``all-or-nothing'' strategy leads to significant computational waste, as the small model often possesses the competence to handle the majority of the reasoning steps, requiring expert intervention only at specific critical positions~\citep{lin2024critical,ruan2025enhancing}.

\begin{figure}
    \centering
    \includegraphics[width=\linewidth]{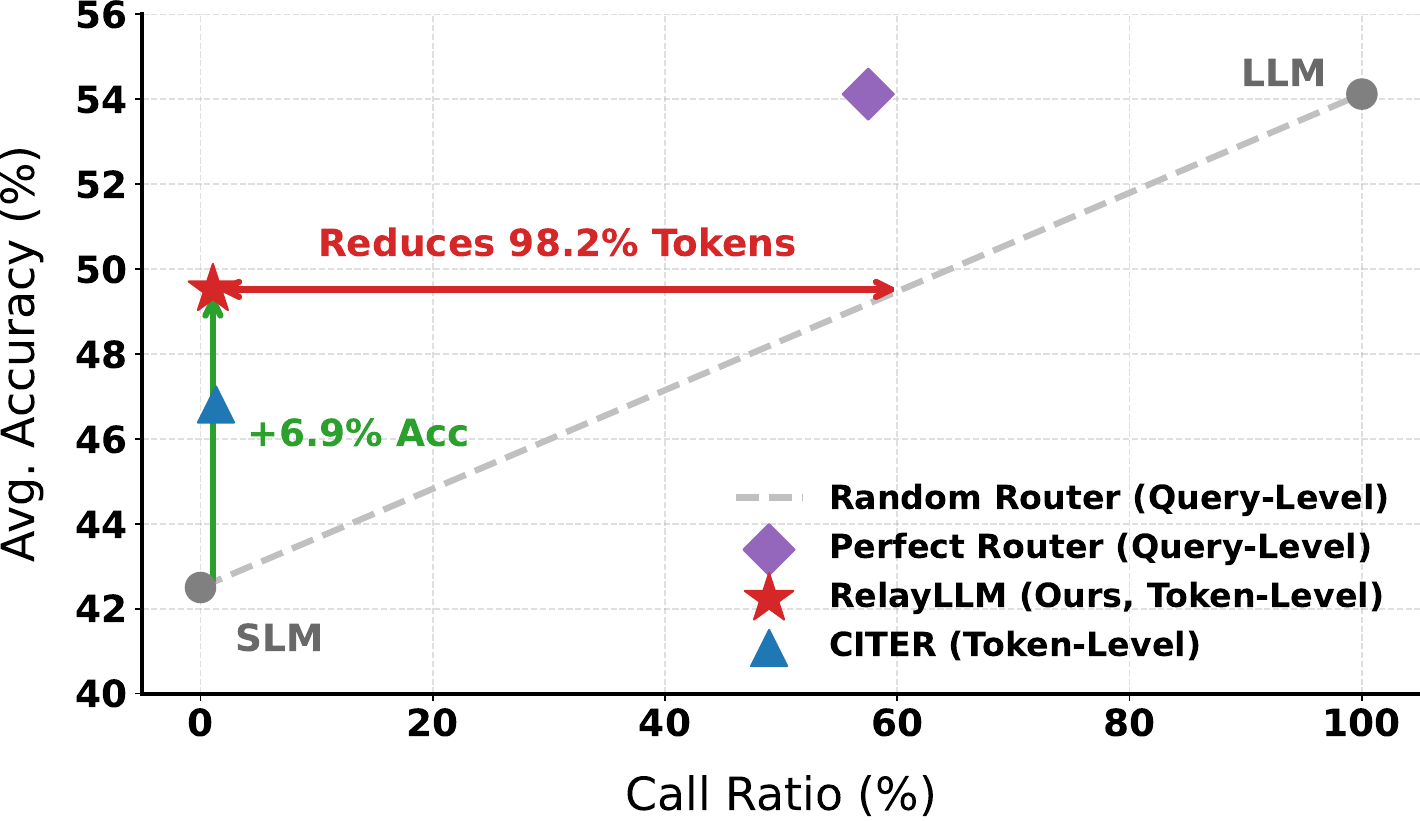}
    \caption{Results are averaged across six mathematical benchmarks. 
The ``Random Router'' baseline randomly directs questions to either the small or large model. 
The ``Perfect Router'' baseline directs only questions SLM cannot solve to large model. 
The x-axis represents the Call Ratio (percentage of tokens generated by the teacher model), and the y-axis denotes the average accuracy.}
    \label{fig:intro}
\end{figure}

To address these problems, we propose \method, a novel framework for efficient reasoning via token-level collaborative decoding~\citep{shen2024learning} without an additional controller. Unlike static routers, \method empowers the small model to act as both a problem solver and an active controller that dynamically requests assistance only when necessary. 

Inspired by tool-use agents~\citep{wolflein2025llm, zheng2025parallel}, we introduce an interleaved generation process where the small model generates a special command token ($\texttt{<call>}$) to pause its own generation and invoke the large model for a specified number of tokens. The small model then receives the expert's guidance and resumes reasoning, effectively ``relaying'' the output process between models. 
We propose a two-stage framework to equip the small model with this strategic delegation capability to train the \method. We first employ a supervised warm-up phase to teach the model the syntactic structure of calling commands. This is followed by a reinforcement learning stage using Group Relative Policy Optimization (GRPO)~\citep{shao2024deepseekmath} training, where we design a context-aware reward that guides the model to balance independence with necessary help-seeking, penalizing both wasted costs and avoidable errors.

Empirical results on six benchmarks demonstrate the effectiveness of our approach. As illustrated in Figure~\ref{fig:intro}, \method~achieves an average accuracy of 49.52\% on six benchmarks, largely recovering the performance gap between the small model and the large one. Remarkably, this gain is achieved with minimal cost, as \method~invokes the large model for only 1.07\% of the total generated tokens. 
Furthermore, in comparison to a resource-equivalent Random Router, \method~yields a substantial 6.9\% accuracy improvement. These results confirm that \method~effectively identifies critical reasoning steps for expert intervention, reducing token costs by 98.2\% compared to a performance-matched router. Surprisingly, evaluations in a teacher-free setting reveal that the model internalizes effective reasoning patterns during collaboration, enabling it to surpass baselines on easier benchmarks even without expert assistance.




\section{\method~Inference}
\label{subsec:framework}
\begin{figure*}
    \centering
    \includegraphics[width=0.98\linewidth]{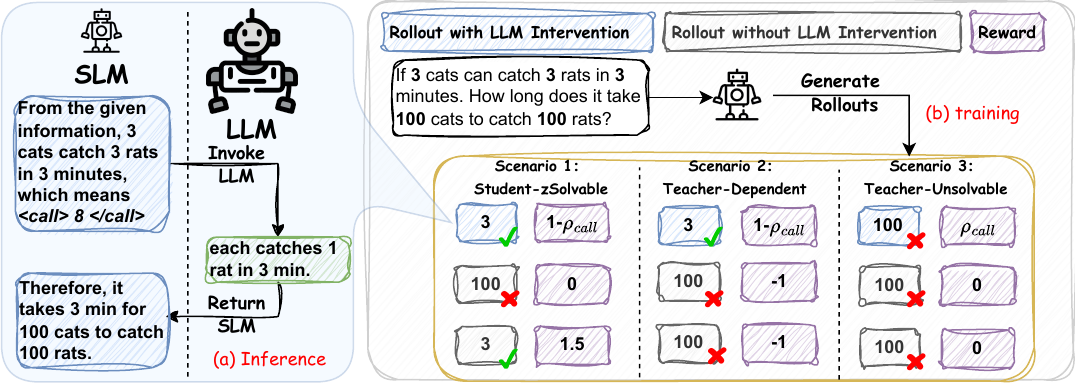}
    \caption{Overview of the \method~ framework.\textbf{(Left) Collaborative Inference:} The Small Language Model acts as a central controller. During generation, it can actively trigger an ``intervention'' by generating a \texttt{<call>} command. The Large Language Model then generates the specified number of tokens, after which control returns to the SLM to complete the reasoning. \textbf{(Right) Difficulty-Aware Reward Design:} During GRPO training, we sample a group of rollouts (both with and without LLM intervention) and classify the query difficulty into three distinct scenarios: \textit{Solvable}, \textit{Teacher-Dependent}, and \textit{Unsolvable}, to guide the policy optimization.Distinct reward designs are applied to each scenario to align the model's behavior with the optimal strategy.}
    \label{fig:framework}
\end{figure*}
As illustrated in Fig.~\ref{fig:framework}, we consider a hybrid inference setting involving a primary, resource-efficient Small Language Model (SLM), denoted as $\mathcal{M}_S$, and a powerful but computationally expensive Large Language Model (LLM), denoted as $\mathcal{M}_L$. Given an input query $\mathbf{x}$, the system aims to generate a high-quality response $\mathbf{y}$ through dynamic collaboration.

Different from standard cascading methods that simply offload the entire remaining task to a larger model, we define an interleaved generation process where the small model acts as both reasoner and central controller. The process operates as follows:

\subsection{Small Model Generation}
By default, $\mathcal{M}_S$ generates tokens autoregressively based on the current context history as a normal language model. $\mathcal{M}_S$ is augmented with a special control capability: it can actively request assistance by generating a specific command pattern:
$$
    \mathcal{C}_{\text{cmd}}(n) = \texttt{<call>} \oplus n \oplus \texttt{</call>}
$$
where $\oplus$ denotes string concatenation, and $n \in \mathbb{Z}^+$ represents the number of tokens required from the larger language model.

\subsection{Large Model Intervention}
When this trigger pattern is detected, the generation by $\mathcal{M}_S$ pauses. The current context (including query and tokens generated by SLM) is forwarded to $\mathcal{M}_L$. 
Crucially, to maintain compatibility with the large model's standard input distribution, we strip the special command tokens ($\mathcal{C}_{\text{cmd}}$) from the context provided to $\mathcal{M}_L$. The large model then generates the next $n$ tokens (or stops early if an end-of-sequence ``[EOS]'' token is reached), providing a high-quality continuation or reasoning step.

\subsection{Iterative Relay of LLM and SLM}
After $\mathcal{M}_L$ completes the generation, control is returned to the small model. The context is updated with the new tokens generated by $\mathcal{M}_L$ added into the existing context. 
Unlike the large model, $\mathcal{M}_S$ retains the full history, including its own generated command tokens (<call>n</call>). This allows $\mathcal{M}_S$ to maintain a trace of its active delegation decisions. With the updated context, $\mathcal{M}_S$ resumes generation, digesting the expert's guidance to continue the reasoning process.


\section{\method~Training}
\label{subsec:training}

To enable the small model $\mathcal{M}_S$ to determine \textit{when} and \textit{how long} to invoke the LLM, we propose a two-stage framework including a supervised warm-up phase followed by a reinforcement learning phase.

\subsection{Cold Start via Supervised Warm-up}
The small model $\mathcal{M}_S$ may not naturally generate the command pattern $\mathcal{C}_{\text{cmd}}$ if we train it directly with reinforcement learning. We employ a supervised warm-up phase as cold start to initialize the model with the necessary structural knowledge.
We construct a synthetic dataset $\mathcal{D}_{\text{warm}}$ to teach the model the calling command without inducing distribution shifts.

To prevent distribution shifts of the small model $\mathcal{M}_S$, 
we first generate base sequences $\mathbf{y}$ by sampling directly from vanilla $\mathcal{M}_S$ rather than relying on external corpora. This ensures that the training contexts align perfectly with the small model's own distribution. Within these self-sampled sequences, we insert command tokens at random indices $t$ at the token level. By avoiding rigid restrictions to sentence or paragraph boundaries, we enable the model to learn to trigger assistance at the precise moment a reasoning gap occurs during the inference time.
This data construction strategy ensures effective on-policy training data to prevent data distribution shift of the small model, while simulating various calling scenarios.

Furthermore, we explicitly simulate varying degrees of reliance on the expert model by synthesizing delegation lengths across multiple orders of magnitude. We randomly sample the target length $n_{\text{sample}} = d \times 10^k$, where $d \in \{1, \dots, 9\}$ and $k \in \{0, \dots, 3\}$. To guarantee validity, the final request length is clipped to the available response via $n = \min(n_{\text{sample}}, L_{\text{rem}})$.

We fine-tune $\mathcal{M}_S$ on this constructed dataset using standard cross-entropy loss. While we acknowledge a theoretical discrepancy between this training setup (where tokens subsequent to the control command are from model itself) and the inference phase (where tokens are generated by $\mathcal{M}_L$), this supervised warm-up is essential. It focuses primarily on adapting the model to output valid executable commands ($\mathcal{C}_{\text{cmd}}$). This creates a stable starting point for the subsequent reinforcement learning stage, where $\mathcal{M}_S$ will be trained to request token-level assistance from $\mathcal{M}_L$ and take advantage of their feedback.
\subsection{Policy Refinement with Reinforcement Learning}
In this stage, we employ Reinforcement Learning to align the model's behavior with our core objective: maximizing response quality while strictly minimizing collaborative cost. We proceed by detailing our optimization framework.

\subsubsection{GRPO Training with RLVR}

We leverage Group Relative Policy Optimization (GRPO)~\citep{shao2024deepseekmath} to refine the policy $\pi_\theta$ of $\mathcal{M}_S$ by adopting the Reinforcement Learning with Verifiable Reward (RLVR) paradigm~\citep{Lambert2024TLU3P}. 
This paradigm is well-suited for domains where the quality of generated responses can be deterministically verified. 
In its standard form, RLVR employs a rule-based verifier $v : \mathcal{X} \to \{0,1\}$ that assigns binary rewards.
This binary scheme is effective for tasks with unambiguous success criteria, such as mathematical problem-solving. In our setting, we design a corresponding rule-based reward to verify model-generated relations, described in Sec.~\ref{sec:reward}.

To optimize the policy using these rewards, GRPO samples a group of outputs $\{o_i\}_{i=1}^G$ for each query $q$ from the old policy $\pi_{\theta_{\mathrm{old}}}$ and evaluates them relative to the group average. The training objective is formulated as:
\begin{equation}
    \mathcal{J}_{\mathrm{GRPO}}(\theta) = \mathbb{E}_{q \sim \mathcal{D}} \left[ \frac{1}{G} \sum_{i=1}^G \left( \mathcal{M}_i - \beta \mathbb{D}_{\mathrm{KL}} \right) \right],
\end{equation}
where $\mathbb{D}_{\mathrm{KL}} = D_{\mathrm{KL}}(\pi_{\theta} \parallel \pi_{\mathrm{ref}})$ is the regularization term. The surrogate objective $\mathcal{M}_i$ is computed as $\min(\rho_i A_i, \mathrm{clip}(\rho_i, 1-\epsilon, 1+\epsilon)A_i)$, where $\rho_i = \frac{\pi_{\theta}(o_i|q)}{\pi_{\theta_{\mathrm{old}}}(o_i|q)}$. The advantage $A_i$ is derived from the group-normalized rewards defined below:
\begin{equation}
    A_i = \frac{r_i - \mathrm{mean}(\{r_j\})}{\mathrm{std}(\{r_j\}) + \varepsilon_{\mathrm{stab}}},
\end{equation}
where $\varepsilon_{\mathrm{stab}}$ is a small constant for stability. This formulation encourages the model to generate responses that outperform the group average.

\subsubsection{Data Filtering}
\label{sec:data_filter}
Since our method leverages the large model to generate reasoning paths or feedback, it is essential to identify the subset of data where such intervention is useful.
If the large model consistently fails to solve a query, calling it during training yields no positive gain.
Therefore, we preprocess the dataset to filter out instances that are too hard for the large model.
We sample 10 responses per query and only preserve those with a pass rate of $\ge 50\%$.
This step ensures that the training data lies in the competence boundary of the large model and the responses can contribute effectively.
We provide an ablation study on this filtering mechanism in  Table~\ref{tab:ablation}.

\subsubsection{Reward Design}
\label{sec:reward}
We formulate the optimization objective using two distinct reward signals, a simple reward and our designed difficulty-aware reward. Let $y$ be the response, $a$ be the final answer parsed from $y$, $g$ be the ground truth, and $\rho(y)\in[0,1]$ be the \textit{call ratio} (the ratio of large-model generated tokens to the total response length.). 

\paragraph{Simple Reward.} We define a straightforward reward to encourage both accuracy and efficiency:
\begin{equation}\label{eq:simple_reward}
r_{\text{simple}}(y) = \mathbb{1}(a=g) - \rho(y).
\end{equation}
where $\mathbb{1}(\cdot)$ is the indicator function, thus the responses are scored by their correctness and penalized by the cost of calling the expert model.

\paragraph{Difficulty-Aware Reward.} To capture the relative difficulty of each query, we define the reward based on the collective performance of the sampled group $\mathcal{G}$. We categorize each query into three scenarios based on its difficulty (correctness of responses in $\mathcal{G}$).
As illustrated in Figure~\ref{fig:framework} (Right), we provide concrete examples of how these rewards are assigned across different categories.
\paragraph{Scenario 1: Student-Solvable (Encouraging Independence).}
This scenario applies when the student model is capable of solving the query independently, without assistance from the large model. This scenario is identified if there exists at least one sample in the group $\mathcal{G}$ that answers correctly \textit{without} invoking the large model.
In this case, calling the teacher is deemed redundant.
Consequently, we assign a boosted bonus ($r=1.5$) for independent success to promote efficiency and independence, while dependent success ($\rho(y)>0$) still receives the simple reward $r_{\text{simple}}$ in Eq.\eqref{eq:simple_reward}, and incorrect responses receive zero reward.

\paragraph{Scenario 2: Teacher-Dependent (Penalizing Stubbornness).}
This scenario represents challenging queries where correct answers appear \textit{only} in samples that call the large model.
Here, the small model's independent reasoning is insufficient.
To discourage blind guessing, we impose a penalty  on samples that fail to call the teacher ($r=-1.0$ when $\rho(y)=0$).
Conversely, effective expert calling that leads to a correct answer is rewarded with $r_{\text{simple}}$.

\paragraph{Scenario 3: Teacher-Unsolvable (Incentivizing Exploration).}
This scenario occurs when no sample in $\mathcal{G}$ yields the correct answer, indicating that the query is extremely difficult or the teacher's guidance was ineffective.
Rather than providing zero  training signal for all responses, we assign a small exploration reward ($r=\rho(y)$) to samples that attempted to call the large model. This reinforces the tendency to seek help in highly uncertain situations.
This piecewise design aligns the policy with an optimal strategy: \textit{solve independently when possible, seek help when necessary, and avoid costly errors.}

\section{Experiments}
\label{sec:experiments}
\subsection{Experimental Setup}

\begin{table*}[htbp]
\centering
\caption{Performance comparison on six benchmarks. We compare the effectiveness of \method~using Qwen3-0.6B and Qwen3-1.7B as student models across different methods: the standard Base model, the GRPO-tuning baseline, CITER and \method~(Simple-Reward and Difficulty-Aware-Reward). The Qwen3-8B teacher model performance is provided for reference. We report avg@32 for the challenging AIME datasets and standard pass@1 (greedy decoding) for all other benchmarks. The ``Avg. Call Ratio'' denotes the percentage of tokens generated by the teacher model during the collaborative inference process. The best results within each model group are highlighted in bold.}
\label{tab:math_benchmark}
\resizebox{\textwidth}{!}{
\begin{tabular}{lcccccccc}
\toprule
\textbf{Model} & \textbf{Minerva} & \textbf{MATH500} & \textbf{GSM8K} & \textbf{Olympiad} & \textbf{AIME25} & \textbf{AIME24} & \textbf{Average} & \textbf{Avg. Call Ratio} \\
\midrule
\multicolumn{9}{l}{\textit{Qwen3-0.6B}} \\
Base Model                 & 15.81 & 54.00 & 64.82 & 26.22 & 1.04 & 1.15 & 27.17 & -- \\
GRPO                       & 17.65 & 58.60 & 65.50 & 29.04 & 5.42 & 3.23 & 29.91 & -- \\
CITER                      & 19.29 & 58.80 & 67.78 & 29.60 & 5.93 & 3.24 & 30.77 & 0.98\% \\
\method~(Simple)    & 20.96 & \textbf{60.20} & 69.14 & 32.15 & \textbf{7.19} & \textbf{3.85} & 32.25 & 0.31\% \\
\method~(Difficulty-Aware)     & \textbf{23.53} & 60.00 & \textbf{71.95} & \textbf{32.74} & 6.15 & \textbf{3.85} & \textbf{33.04} & 0.77\% \\
\midrule
\multicolumn{9}{l}{\textit{Qwen3-1.7B}} \\
Base Model                 & 33.82 & 74.60 & 82.64 & 43.11 & 8.75 & 12.08 & 42.50 & -- \\
GRPO                       & 35.66 & 75.60 & 81.73 & 45.04 & 10.73 & 15.62 & 44.06 & -- \\
CITER                      & 38.63 & 80.24 & 82.26 & 51.20 & 11.96 & 16.58 & 46.81 & 1.34\% \\
\method~(Simple)    & 43.01 & \textbf{83.40} & 86.13 & 51.56 & \textbf{13.44} & \textbf{18.23} & 49.30 & 0.43\% \\
\method~(Difficulty-Aware)     & \textbf{43.75} & 81.40 & \textbf{86.28} & \textbf{55.70} & 12.71 & 17.29 & \textbf{49.52} & 1.07\% \\
\midrule\midrule
\rowcolor{gray!20}Qwen3-8B & 48.16 & 83.20 & 93.63 & 56.89 & 17.92 & 24.90 & 54.12 & 100\% \\
\bottomrule
\end{tabular}}
\end{table*}

\subsubsection{Models}
To evaluate the effectiveness of \method, we utilize the Qwen3 model family \citep{yang2025qwen3} due to its consistent architectural scaling and strong performance across various sizes. We select Qwen3-0.6B and Qwen3-1.7B as our primary small language models ($\mathcal{M}_S$) to investigate how our framework scales with model capacity at the sub-2B parameter level.
For the teacher model ($\mathcal{M}_L$), we utilize Qwen3-8B. Selecting a model from the same model family ensures that the generation style, token distribution, vocabulary and tokenizer are more consistent, making collaboration more stable. To optimize training and inference efficiency, we consistently run the models in non-thinking mode.

\subsection{Evaluation Setup}

To evaluate the effectiveness of ~\method, we conduct experiments on six reasoning benchmarks, and compare our approach against the standard GRPO baseline. We also add CITER~\citep{zheng2025citer}, a token-level routing method as the baseline method which requires an additional controller.
The benchmarks include Minerva \citep{Lewkowycz2022SolvingQR}, MATH-500 \citep{Hendrycks2020MeasuringMM}, GSM8K \citep{Cobbe2021TrainingVT}, Olympiad-Bench \citep{He2024OlympiadBenchAC}, AIME-2024, and AIME-2025. We use GPT-4o-mini as a semantic judge to verify the model's output against the ground truth \citep{Zhao2025OneTT}. For the high-difficulty AIME datasets, we report the avg@32 metric to ensure a robust evaluation. For other benchmarks, we report standard accuracy (pass@1) using greedy decoding.

\subsection{Training Details}
We conduct our experiments using the DAPO dataset~\citep{yu2025dapo}.
Our implementation is built upon the EasyR1 framework~\citep{zheng2025easyr1} using its default hyperparameter configurations (shown in App.~\ref{app:hyper}).
All models are trained for a single epoch to ensure a fair comparison.
Regarding data usage, the GRPO baseline is trained on the full dataset, whereas our method utilizes the filtered subset as described in Section~\ref{sec:data_filter}.
To enable efficient interaction with the large model, we serve the teacher model via the vLLM inference engine~\citep{kwon2023efficient}.
We implement the switching mechanism as a stop sequence in the sampling parameters: when the model generates the calling command token, generation halts, and the system invokes the teacher model via the API.

\subsection{Main Results}

As presented in Table \ref{tab:math_benchmark}, \method demonstrates a superior trade-off between reasoning capability and inference efficiency. 
First, our method achieves substantial performance improvements across all benchmarks while maintaining a negligible collaborative cost (less than 1\% token overhead). 
For instance, on the challenging Minerva benchmark, the Qwen3-0.6B model with Difficulty-Aware-Reward improves from a base score of 15.81\% to 23.53\%, representing a relative improvement of approximately 48.8\% while invoking the large model for only 0.77\% of the total tokens. Compared to CITER, our method demonstrates superior performance despite CITER's more computationally intensive design. CITER relies on an external MLP to estimate a score every token, which introduces substantial latency and computational overhead. In contrast, \method~achieves better results with a significantly more efficient mechanism at the cost of only several addtional tokens.

Second, comparing optimization strategies, the Difficulty-Aware-Reward mechanism outperforms the Simple-Reward in performance, with a marginal increase in token consumption. 
For the Qwen3-1.7B model, the Difficulty-Aware-Reward strategy achieves a higher average accuracy of 49.52\% compared to 49.30\% for the Simple-Reward, which correlates with a slight increase in the average call ratio from 0.43\% to 1.07\%, suggesting that the difficulty-based signal better incentivizes the model to seek help in complex scenarios. 

Finally, ~\method~ effectively bridges the capability gap between small and large models using minimal tokens. 
Remarkably, the Qwen3-1.7B (Difficulty-Aware) recovers approximately 60\% of the performance gap between the base SLM (42.50\%) and the expert model (54.12\%), highlighting that sparse, strategic intervention at critical reasoning steps is sufficient to unlock a significant portion of the teacher model's potential.

\section{Analysis}
In this section, we conduct a series of in-depth analyzes to better understand the behavior and effectiveness of \method~framework.

\subsection{\method Generalizes to Unseen Reasoning Domains}
To verify the generalization capability of \method, we extended our evaluation to general reasoning domains that were unseen during training. 
Although our model was trained exclusively on the mathematical DAPO dataset, we tested it on three diverse benchmarks out of the math domain: Big-Bench Hard (BBEH)~\citep{kazemi2025big}, MMLU-Pro~\citep{wang2024mmlu}, and SuperGPQA~\citep{du2025supergpqa}. 
As shown in Table \ref{tab:benchmark_results}, \method~consistently outperforms baseline methods despite the domain shift.
For instance, using Qwen3-1.7B, our method achieves 59.03\% on MMLU-Pro, significantly surpassing the GRPO baseline (49.76\%) and CITER (53.38\%).
These results demonstrate that our framework effectively help SLM have a generalized help-seeking behavior; even when facing unfamiliar inputs, the SLM successfully recognizes its knowledge gaps and invoke the LLM, leading to substantial performance gains in out-of-distribution tasks.
\begin{table}[t]
\centering
\caption{Performance comparison on reasoning and general knowledge benchmarks. We evaluate the effectiveness of \method~using Qwen3-0.6B and Qwen3-1.7B as student models across different settings. The Qwen3-8B performance is provided for reference. The best results within each model group are highlighted in bold.}
\label{tab:benchmark_results}
\resizebox{0.48\textwidth}{!}{
\begin{tabular}{lccc}
\toprule
\textbf{Model} & \textbf{BBEH} & \textbf{MMLU-Pro} & \textbf{SuperGPQA} \\
\midrule
\multicolumn{4}{l}{\textit{Qwen3-0.6B}} \\
Base Model              & 7.19 & 30.03 & 17.22 \\
GRPO                    & 7.82 & 32.15 & 19.91 \\
CITER                   & 8.16 & 33.12 & 20.34 \\
\method~(Simple) & 8.32 & 35.61 & \textbf{21.35} \\
\method~(Difficulty-Aware)  & \textbf{8.56} & \textbf{35.87} & 20.88 \\
\midrule
\multicolumn{4}{l}{\textit{Qwen3-1.7B}} \\
Base Model              & 9.91 & 46.90 & 24.46 \\
GRPO                    & 10.89 & 49.76 & 26.01 \\
CITER                   & 11.67 & 53.38 & 28.25 \\
\method~(Simple) & \textbf{12.67} & 58.76 & 29.85 \\
\method~(Difficulty-Aware)  & 12.46 & \textbf{59.03} & \textbf{29.93} \\
\midrule\midrule
\rowcolor{gray!20}Qwen3-8B & 15.31 & 66.46 & 36.21 \\
\bottomrule
\end{tabular}}
\end{table}

\subsection{Ablation Study}

To investigate the distinct contribution of each component in \method~, we conducted an ablation study using the Qwen3-1.7B model in Table~\ref{tab:ablation}.

\paragraph{Data filtering prevents wasteful calls where teacher models fail.}
We show that removing the data filtering mechanism results in a tripled call ratio with decreased accuracy; this confirms that filtering out queries intractable for the teacher is crucial to avoid cost that yield no performance gain. Filtering out some too hard data also save time and resources during the training stage.

\paragraph{Encouraging independence reduces reliance on teacher model and improves efficiency.}
We remove the independence encouraging (where we boosted correctness reward from $1$ to $1.5$ for solvable queries), and this causes the call ratio to spike to 4.10\%. This demonstrates that specifically rewarding the independent success is crucial to prevent the model from becoming over-reliant on the expert LLM for tasks it could solve alone. 

\paragraph{Exploration reward effectively increases accuracy.}
When we remove the exploration reward (for unsolvable queries), this leads to a significant accuracy drop to 47.56\%, indicating that the exploration reward is necessary to encourage the model to call for help from teacher models in highly uncertain scenarios.
\begin{table}[!t]
    \centering
    \small
    \caption{Ablation study on data filtering and reward design strategies using Qwen3-1.7B. 
    ``w/o Data Filtering'' denotes training on the unfiltered dataset including teacher-failed queries. 
    ``w/o Indep. Incentive'' removes the bonus reward (from $1.5$ to $1$) for independent success (Scenario 1). 
    ``w/o Explor. Reward'' removes the exploration reward (from $\rho$ to $0$) for seeking help in unsolvable queries (Scenario 3).}
    \begin{tabular}{lcc}
        \toprule
        \textbf{Method} & \textbf{Avg. Acc. (\%)} & \textbf{Call Ratio (\%)} \\
        \midrule
        \textbf{\method} & \textbf{49.52} & 1.07 \\
        \quad w/o Data Filtering & 48.76 & 3.30 \\
        \quad w/o Indep. Incentive & 49.34 & 4.10 \\
        \quad w/o Explor. Reward & 47.56 & 0.65 \\
        \bottomrule
    \end{tabular}
    \label{tab:ablation}
\end{table}
\subsection{Intrinsic Reasoning Capability}
\label{sec:intri}
To investigate whether ~\method~ improves the student's inherent reasoning or merely learns to offload tasks, we evaluate the models in a ``Teacher-Free'' setting by strictly forbidding invocations during inference (implemented via \texttt{bad\_words=[``<call>'', ``</call>'']} when inference). 
Results in Table \ref{tab:intrinsic} reveal three key insights.
First, on Easy datasets, even without teacher access, ~\method~ (Simple-Reward) achieves 61.12\%, surpassing the GRPO baseline. This suggests that the student model has successfully learn  from the  reasoning ability of the expert model during the collaborative training process. 

On Harder datasets, removing the teacher leads to a notable performance drop (e.g., Difficulty-Aware-Reward falls from 15.00\% to 11.93\%), confirming that for complex tasks, the model remains heavily dependent on expert intervention. 
Third, comparing reward schemes, the Simple-Reward variant demonstrates stronger capabilities than the Difficulty-Aware-Reward variant. This aligns with our previous observation that Difficulty-Aware-Reward encourages a higher call ratio, leading to a stronger dependency on the teacher, whereas Simple-Reward retains more independence.
\begin{table}[t]
    \centering
    \small
    \caption{Evaluation of intrinsic reasoning capability. We disable the teacher during inference (``w/o Teacher'') by masking the call tokens. ``Hard'' refers to AIME24 and AIME25, while ``Easy'' refers to the remaining.}
    \begin{tabular}{lcc}
        \toprule
        Method & Easy (\%) & Hard (\%) \\
        \midrule
        GRPO Baseline & 59.51 & 13.18 \\
        \midrule
        \multicolumn{3}{l}{\textit{~\method~ (Simple)}} \\
        \quad Standard Inference & 66.03 & 15.84 \\
        \quad w/o Teacher & 61.12 & 13.13 \\
        \midrule
        \multicolumn{3}{l}{\textit{~\method~ (Difficulty-Aware)}} \\
        \quad Standard Inference & 66.78 & 15.00 \\
        \quad w/o Teacher & 60.26 & 11.93 \\
        \bottomrule
    \end{tabular}
    \label{tab:intrinsic}
\end{table}
\begin{table}[t]
    \centering
    \small
    \caption{Comparison between dynamic length prediction (\method) and fixed delegation length strategies. 
    Note that ``Fixed-$k$'' does not simply denote inference truncation; these models were \textbf{retrained} with the constraint to always request $k$ tokens, ensuring they learned optimal policies for those specific lengths.}
    \begin{tabular}{lcc}
        \toprule
        \textbf{Method} & \textbf{Avg. Acc. (\%)} & \textbf{Call Ratio (\%)} \\
        \midrule
        Fixed-20 & 49.41 & 1.32 \\
        Fixed-100 & 49.56 & 2.87 \\
        Fixed-500 & \textbf{51.17} & 5.37 \\
        \midrule
        \textbf{~\method~} & 49.52 & \textbf{1.07} \\
        \bottomrule
    \end{tabular}
    \label{tab:length_ablation}
\end{table}

\subsection{Dynamic Token-Length Calling Minimizes Computational Cost}
We investigate whether dynamically predicting the calling length $n$ requested from the large model is superior to using rigid, pre-defined lengths. 
To ensure a fair evaluation, we \textbf{retrained} separate variations of the student model where the call command is hard-constrained to a fixed token count $k \in \{20, 100, 500\}$ during both training and inference.
As shown in Table \ref{tab:length_ablation}, ~\method~ demonstrates superior efficiency compared to these specialized fixed-length models.
Specifically, compared to the \textbf{Fixed-100} model, ~\method~ achieves a similar accuracy but reduces the call ratio from 2.87\% to 1.07\%.
This indicates that while the fixed-length model is forced to consume a set budget even for simple queries, ~\method~ effectively learns to request ``just enough'' tokens to bridge the reasoning gap, thereby minimizing computational waste without compromising performance. We provide a more detailed results in Appendix~\ref{app:length_ablation}.

\begin{figure}[t]
    \centering
    \includegraphics[width=\linewidth]{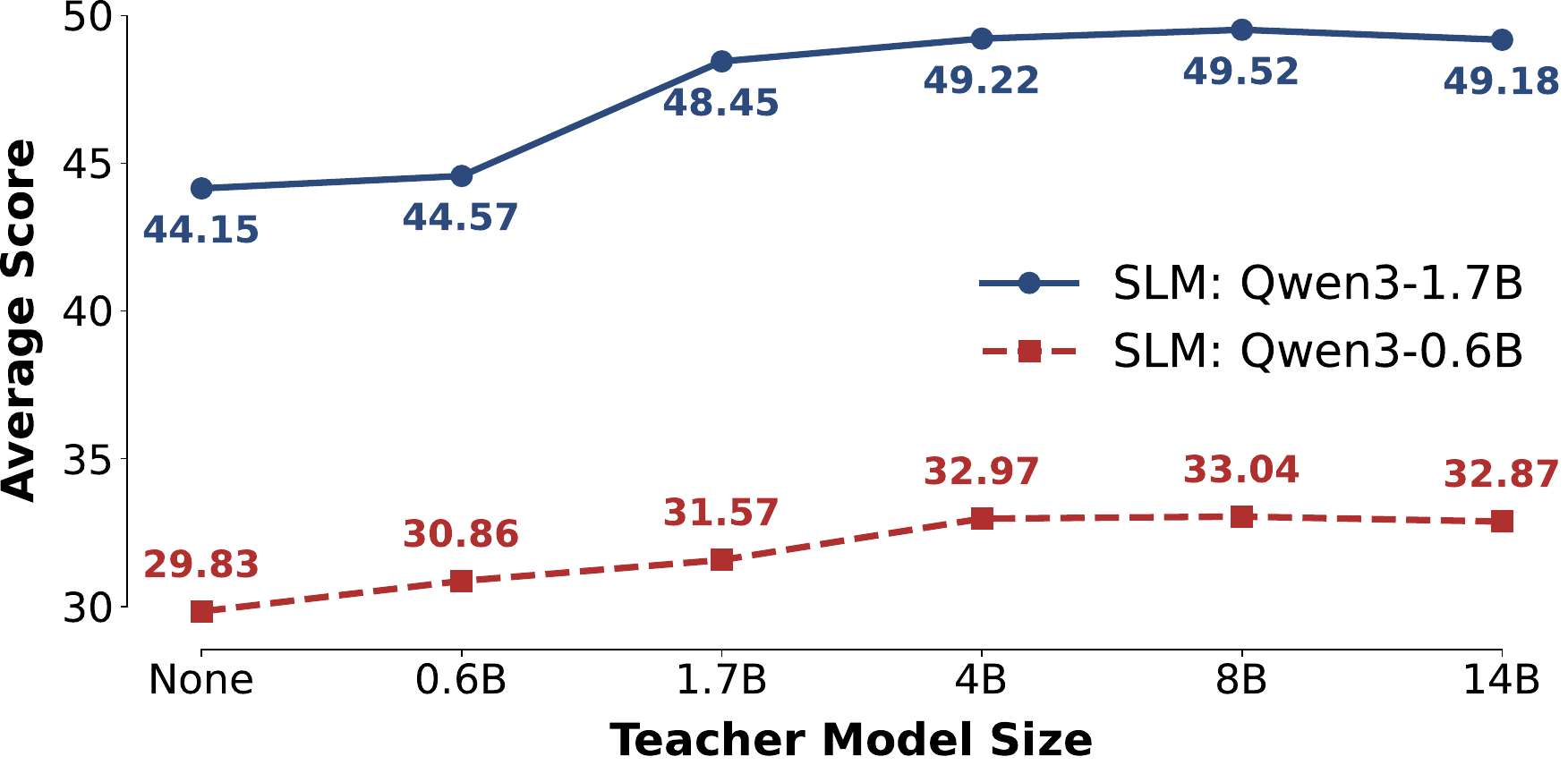}
    \caption{Impact of teacher model size on student performance. We evaluate two student models (Difficulty-Aware) across six benchmarks. The x-axis represents the size of the teacher model used during the collaborative inference process, ranging from "None" (the same as Sec.~\ref{sec:intri}, prevent model to call teacher model) to 14B. The reported scores are averaged across all six datasets.}
    \label{fig:teacher_size}
\end{figure}
\subsection{Distributional Alignment }
To determine whether the student model has acquired generalized reasoning capabilities or merely overfitted to the specific patterns and words of the LLM in training time, we performed a cross-LLM evaluation by substituting the training teacher with different models in the inference phase.
The results, illustrated in Figure \ref{fig:teacher_size}, reveal two critical insights.
First, consistency with the training LLM yields optimal performance. 
The accuracy peaks when the inference teacher matches the training teacher, reaching 49.52\% for the 1.7B student. 
Notably, replacing it with a larger model results in a slight performance decline. 
This indicates that the distribution shift between the training and inference teachers can outweigh the benefits of the larger model's superior reasoning capabilities.

Second, even employing a relatively weak teacher that is weaker than itself (e.g., 0.6B or 1.7B) consistently outperforms the ``None'' baseline. This suggests that the trained model has become accustomed to the presence of external assistance, adapting its generation dynamics to effectively leverage such interventions rather than relying solely on its intrinsic capabilities.
Furthermore, excluding the distribution shift at 8B, which is we used in training, there is a positive correlation between teacher size and student performance, confirming that the student effectively leverages the stronger reasoning signals provided by more capable experts.


\section{Related Work}

\subsection{Model Collaboration}
Model collaboration~\citep{feng2025one} ranges from weight-level merging~\citep{wortsman2022model, huang2023lorahub} and logits-level ensembling~\citep{liu2024tuning,li2023contrastive} to text-level interaction. Recent research has focused on navigating the efficiency trade-off between large and small models, which can be broadly categorized into two directions.
The first direction involves speculative reasoning. \citet{bachmann2025judge} and \citet{pan2025specreason} employ judge mechanisms or verifiers to validate small model outputs, effectively acting as dynamic routers. Similarly, large model guidance is leveraged to enhance small model reasoning specifically at inference time \citep{yang2025speculative,zhang2025router}. 
The second direction focuses on collaborative decoding~\citep{shen2024learning} with interleaved generation via an additional controller. \citet{shen2024learning,sun2024adaswitch} proposed learning a joint policy for multiple models. Strategic intervention is further investigated by \citet{li2025efficient,feng2025don,fu2025r2r} to explore thought spaces efficiently.

\subsection{RL for LLM Reasoning}
Reinforcement learning has recently emerged as a pivotal technique for augmenting LLM reasoning, demonstrating broad success ranging from traditional mathematical and code generation tasks~\citep{guo2025deepseek,wang2025code} to intricate multi-modal challenges~\citep{huang2025vision, wang2025vl,li2025self} and structured data environments~\citep{shi2025mobilegui, tang2025unirelr1rltunedllmreasoning}. To support these diverse applications and enable complex behaviors like ~\method, concurrent research is actively refining methodologies through novel training paradigms, such as self-play~\citep{liu2025spiral,huang2025r,yu2025guided}, alongside developing more robust algorithmic techniques exemplified by DAPO~\citep{yu2025dapo}, VAPO~\citep{yue2025vapo}, and high-entropy guided optimization~\citep{dai2025cde,wang2025beyond,Zhou2025EvolvingLM}.



\section{Conclusion}
We presented \method, addressing the inefficiency of "all-or-nothing" offloading in routing systems. By treating the large model as an on-demand tool rather than a fallback generator, our approach demonstrates that small models can handle the vast majority of reasoning steps if supported at specific critical positions. The success of our GRPO-based training strategy confirms that help-seeking behaviors can be effectively learned and optimized.
Our results show that \method not only outperforms resource-equivalent random routers by 6.9\% but also achieves comparable reasoning accuracy to larger models with negligible computation.

\section*{Acknowledgments}
We would like to thank Dongfu Jiang (University of Waterloo) for his helpful insights and discussions on tool use LLM.
This research was supported in part by the NVIDIA Academic Grant Program and WashU Ignite Interdisciplinary Grants.

\bibliography{custom}
\clearpage
\appendix

\section{Case Study}
\label{app:case_study}
Here we show two case studies from Qwen3-1.7B (Difficulty-Aware) to illustrate how invoking LLM can help solve the questions in Figure~\ref{fig:case_2}.

\begin{figure*}
    \centering
    \begin{tcolorbox}[
        colback=white,
        colframe=black!75,
        sharp corners,
        enhanced jigsaw,
        breakable,
    fontupper=\small
    ]
    
    \textbf{Case Study:} A list of positive integers has the following properties:\\$\bullet$ The sum of the items in the list is $30$.\\$\bullet$ The unique mode of the list is $9$.$\\\bullet$ The median of the list is a positive integer that does not appear in the list itself.Find the sum of the squares of all the items in the list. \\[3pt]
    
    We are given a list of **positive integers** with the following properties:\\
1. **Sum of the items is 30.**\\
2. **The unique mode is 9.**\\
3. **The median is a positive integer that does not appear in the list.**\\
We are to find the **sum of the squares of all the items** in the list.\\
---     \\                                      \#\#\# Step 1: Understand the properties
- **Mode is 9**: This means 9 appears more times than any other number in the list.
- **Unique mode is 9**: No other number appears as frequently as 9.\\
- **Median is a positive integer that does not appear in the list**: So the median is some number $ m $, and $ m $ is not in the list.\\
- **Sum of the items is 30**.
\\---\\
\#\#\#  Step 2: Consider the structure of the list
Let’s denote the list as $ a_1, a_2, \dots, a_n $, sorted in non-decreasing order.
\#\#\#\# Key observations:
- The **median** is the middle value. If $ n $ is odd, it is the $ \frac{n+1}{2} $-th element. If $ n $ is even, it is the average of the two middle elements.
- Since the **median is a positive integer that does not appear in the list**, it must be **greater than the largest number in the list** (if the list has only one element, it’s trivial, but we'll assume $ n > 1 $).
- The **mode is 9**, so 9 appears more times than any other number.
 \\---\\
\#\#\# Step 
\begin{lstlisting}[basicstyle=\ttfamily\small, breaklines=true, breakatwhitespace=true,escapeinside={(*}{*)}]
(*\color{red}<call> 300 </call>*)
3: Try small values of $ n $
Let's test small values of $ n $ to find a valid list that satisfies all conditions.
#### Try $ n = 5 $ (odd):
- Then the **median is the 3rd element**.
- Suppose the list is $ [a, b, m, c, d] $, where $ m $ is the median.
- $ m $ is not in the list.
- The **mode is 9**, so 9 must appear at least twice, and more than any other number.
Let's try to construct such a list:
Try $ [9, 9, 1, 1, 10] $ (sum = 9+9+1+1+10 = 30)
Check the properties:
- **Sum = 30 **
- **Mode = 9** (appears twice) - but are there other numbers with same frequency?
- 1 appears 3 times -> **mode is 1**, not 9 correct
Try $ [9, 9, 1, 2, 9]$ -> sum = 9+9+1+2+9=30
- Mode is 9 (3 times), 1 and 2 once -> correct
- Median is 1 (the middle number) -> 1 is in the list incorrect
Try $ [9, 9

(*\color{red}(end call)*)
\end{lstlisting}
, 1, 2, 8] \$ -> sum = 9+9+1+2+8 = 29 incorrect
Try $ [9, 9, 1, 3, 8] $ -> sum = 30
- Mode is 9
- Median is 1 (not in the list) 
- 1 is not in the list 
So this is a valid list:
$$
[9, 9, 1, 3, 8]
$$
Check sum: $ 9 + 9 + 1 + 3 + 8 = 30 $ 
Check mode: 9 appears twice, others once -> **unique mode is 9** 
Check median: middle element is 3 → 3 is not in the list 
\\---\\
\#\#\# Step 4: Compute the sum of squares
Compute $ 9^2 + 9^2 + 1^2 + 3^2 + 8^2 $
$
= 81 + 81 + 1 + 9 + 64 = 236
$
\\---\\
$
\boxed{236}
$
    \end{tcolorbox}
  \vspace{-0.1in}
  \caption{A qualitative case study showcases that our model invoke LLM to show how to validate the final answer, then solve the problem by itself.}
  \label{fig:case_2}
  \vspace{-0.1in}
\end{figure*}

\begin{table*}[]
    \centering
    \small
    \caption{Full performance breakdown of different delegation length strategies across six benchmarks. 
    ``Fixed-$k$'' indicates models retrained with a fixed call length of $k$ tokens.
    We report mean@32 for AIME datasets (AIME24, AIME25) and pass@1 for others. 
    Best accuracy values are bolded, and the most efficient call ratio is highlighted.}
    \begin{tabular}{lcccccccc}
        \toprule
        \textbf{Method} & \textbf{Minerva} & \textbf{MATH500} & \textbf{GSM8K} & \textbf{Olympiad} & \textbf{AIME25} & \textbf{AIME24} & \textbf{Average} & \textbf{Call Ratio} \\
        \midrule
        Fixed-20 & 39.71 & 81.40 & 86.96 & 55.74 & 14.11 & 18.54 & 49.41 & 1.32\% \\
        Fixed-100 & 40.44 & 81.40 & 86.50 & 56.78 & 14.42 & 17.81 & 49.56 & 2.87\% \\
        Fixed-500 & \textbf{44.49} & \textbf{81.80} & \textbf{88.25} & \textbf{57.96} & \textbf{15.05} & \textbf{19.48} & \textbf{51.17} & 5.37\% \\
        \midrule
        \textbf{RelayLLM} & 43.75 & 81.40 & 86.28 & 55.70 & 12.71 & 17.29 & 49.52 & \textbf{1.07\%} \\
        \bottomrule
    \end{tabular}
    \label{tab:full_length_ablation}
\end{table*}

\section{Detailed Analysis of Delegation Length Strategies}
\label{app:length_ablation}

In the main text, we demonstrated that RelayLLM achieves comparable average accuracy to a fixed 100-token delegation strategy while consuming significantly fewer tokens. Table \ref{tab:full_length_ablation} presents the granular performance breakdown across all six mathematical benchmarks.

We observe distinct trends across datasets of varying difficulty:

\paragraph{Efficiency on Standard Benchmarks.} 
On standard reasoning datasets such as MATH500 and GSM8K, RelayLLM matches the performance of the Fixed-100 strategy almost exactly (e.g., 81.40\% on MATH500 for both methods). However, it achieves this parity with a drastically lower call ratio (1.07\% vs. 2.87\%). This indicates that for the majority of queries in these datasets, the student model only requires short, precise interventions rather than long-context guidance.

\paragraph{Adaptability on Complex Benchmarks.} 
The advantage of dynamic prediction becomes more evident on the challenging Minerva benchmark. ~\method~ outperforms the Fixed-20 significantly (43.75\% vs. 39.71\%), suggesting that our model correctly identifies the need for longer generation lengths when facing harder problems. While the Fixed-500 strategy yields the highest accuracy on Minerva (44.49\%) and AIME, this marginal gain comes at a prohibitive cost: it incurs a 5.37\% call ratio, which is over $5\times$ the computational overhead of RelayLLM.

\section{Prompt Templates}
This section presents the exact prompt templates used for the models.
\subsection{Inference Prompt}
\begin{tcolorbox}[
  colback=blue!5!white,
  colframe=blue!75!black,
  title=\textbf{Solver Prompt Template},
  fonttitle=\bfseries,
  sharp corners
]
\textbf{System Message:}
\par 
Please reason step by step, and put your final answer within \textbackslash boxed\{\}.

\textbf{User Message:}
\par
\textit{\{problem\_statement\}}
\par
\vspace{0.5em}
\textit{Note: \texttt{\{problem\_statement\}} is a placeholder for the actual math problem.}
\end{tcolorbox}

\subsection{GPT-4o-mini Judge Prompt}
To programmatically evaluate the correctness of answers on mathematical benchmarks where the final answer can be complex (e.g., simplified expressions), we use GPT-4o-mini as a judge. The exact prompt and configuration used for this evaluation are detailed below.

\begin{tcolorbox}[
  colback=blue!5!white,
  colframe=blue!75!black,
  title=\textbf{Configuration for GPT-4o as Judge},
  fonttitle=\bfseries,
  sharp corners
]
\begin{itemize}
    \item \textbf{Model}: \texttt{gpt-4o}
    \item \textbf{Temperature}: 0.1
\end{itemize}

\textbf{System Message:}
\begin{quote}
You are a math answer checker.
\end{quote}

\textbf{User Message Template:}
\begin{quote}
Hi, there is an answer: \texttt{\{answer\}},

and the ground truth answer is: \texttt{\{response\}},

please check whether the answer is correct or not, and return the **only** Yes or No.
\end{quote}
\textit{Note: \texttt{\{answer\}} is a placeholder for the model-generated solution, and \texttt{\{response\}} is the ground-truth answer from the benchmark.}
\end{tcolorbox}

\section{Other Related Work}
~\method~also align with efficient reasoning, include efficient chain-of-thought methods~\citep{wang2025efficient, huang2025efficient}, speculative best-of-N decoding~\citep{sun2024fast}, in-context learning methods~\citep{huang2024divide} and non-myopic generation~\citep{ma2024non}.

\section{Hyperparameter}
\label{app:hyper}

We utilize the Group Relative Policy Optimization (GRPO) algorithm for post-training. The model is optimized using \texttt{AdamW} with a constant learning rate of $1 \times 10^{-6}$ and a weight decay of $1 \times 10^{-2}$. The global batch size is set to 32. 

To support extensive reasoning steps required for complex mathematical problems, we configure the maximum sequence length to allow for long Chain-of-Thought (CoT) generation. Specifically, the maximum prompt length is set to 4096 tokens, and the maximum response length is extended to 8192 tokens. 

For the GRPO specific configurations, we set the group size $G=8$, meaning that 8 outputs are sampled for each prompt to estimate the baseline. To ensure training stability and prevent the policy from deviating excessively from the reference model, we apply a KL divergence coefficient $\beta$ of 0.01. During the rollout phase, we use a sampling temperature of 1.0 to encourage diverse reasoning paths.

\begin{table}[h]
    \centering
    \caption{Hyperparameter settings for GRPO training.}
    \label{tab:hyperparameters}
    \begin{tabular}{lc}
        \toprule
        \textbf{Hyperparameter} & \textbf{Value} \\
        \midrule
        Optimizer & AdamW \\
        Learning Rate & $1 \times 10^{-6}$ \\
        Weight Decay & $1 \times 10^{-2}$ \\
        Global Batch Size & 32 \\
        \midrule
        Max Prompt Length & 4096 \\
        Max Response Length & 8192 \\
        Temperature & 1.0 \\
        \midrule
        Group Size ($G$) & 8 \\
        KL Coefficient ($\beta$) & 0.01 \\
        \bottomrule
    \end{tabular}
\end{table}
\end{document}